\documentclass{article}
\usepackage{spconf,amsmath,graphicx}
\usepackage{enumitem}

\title{Beyond the binary: Limitations and possibilities of gender-related speech technology research}
\twoauthors
  {Ariadna Sanchez$^*$, Alice Ross\sthanks{These authors contributed equally to this work.}}
	{The Centre for Speech Technology Research, \\
        University of Edinburgh, UK}
  {Nina Markl}
	{Institute for Analytics and Data Science, \\ University of Essex, UK}
\begin{document}
%
\maketitle
\begin{abstract}
This paper presents a review of 107 research papers relating to speech and sex or gender in ISCA Interspeech publications between 2013 and 2023. We note the scarcity of work on this topic and find that terminology, particularly the word \textit{gender}, is used in ways that are underspecified and often out of step with the prevailing view in social sciences that gender is socially constructed and is a spectrum as opposed to a binary category. We draw attention to the potential problems that this can cause for already marginalised groups, and suggest some questions for researchers to ask themselves when undertaking work on speech and gender.
\end{abstract}

\section{Introduction}
In contemporary Western science, it has been traditional to categorise humans into two groups -- male and female -- based on genetic, physiological, social and cultural factors or a combination of these. Subversion and rejection of this binary categorisation in society has a long history, and understanding of its limitations in mainstream opinion is now steadily increasing \cite{feinberg1997transgender, hyde2019future}. Recent research draws attention to heterogeneity and fluidity within the groups and to the `grey areas' outside the binary where a rich, varied range of identities exist \cite{sumerau2020tale}. Increasing awareness of non-binary and transgender lived experiences has resulted in changes in legal status and in popular opinion; for example, a 2016 survey of 17,105 adults in 23 countries found that, in every country, the majority supported important transgender rights \cite{flores2016public}. Meanwhile, through analytic frameworks such as Design Justice \cite{costanza2020design} and Queer Data \cite{guyan2022queer}, there has been a vital reckoning of the dangers of exclusion and erasure of marginalised groups when building new technologies, and a growing understanding of the ethical concerns related to this \cite{keyes2018misgendering, markl2022mind}. Using oversimplified generalisations in our models and datasets can reproduce and amplify human biases, with impacts on the resulting technology ranging from consistent underperformance for specific groups of users to being actively harmful.

In the tradition of \cite{moore19_interspeech} and \cite{markl2023everyone}, we examine the frequently underspecified use of the terms \textit{sex} and \textit{gender} in the study of speech, highlighting some inaccuracies and discrepancies. We choose to review papers from ISCA Interspeech for two reasons. First, to show a snapshot of how sex and gender terms are used in gender-related speech technology research more generally. Second, to review a conference with both more technical and interdisciplinary speech research. We invite readers to consider the limitations and potential drawbacks of the lack of clarity and expansiveness in the way these factors are described in contemporary speech science research.

\section{What is gender?}
In line with the consensus view in gender studies, we regard gender as a social construct \cite{butler2002gender, morgenroth2021effects, tannenbaum2019sex}. This does not mean that gender is not `real' or does not have a direct impact on our lives; rather, it means that gender is created by humans. It often entails culturally-specific expectations or stereotypes about behaviour and personality traits, with individuals choosing to conform to these norms to a greater or lesser extent. Psychologically, gender can also refer to a person's individual, internal sense of whether they are a man, a woman, non-binary, agender, gender-fluid, or another identity\def\thefootnote{1}\footnote{https://transactual.org.uk/glossary/}. Linguistically, gender can inform the correct ways to talk to or about a person, e.g., their pronouns. It is perhaps simplest to state what gender is \textit{not}: it is not innate, it is not a binary variable, and it is not possible to determine accurately from a person's face, body, or voice. We can, and often do, guess based on visual cues, but the only way to \textit{know} somebody's gender is to ask them. By contrast, sex generally refers to biological attributes associated with being male, female, or intersex, such as chromosomes and genitals in humans \cite{viloria2020spectrum}. The term `sex assigned at birth' emphasises the fact that, for many people, this does not align with their gender -- and that a person's physical characteristics can be altered through surgery or hormone therapy. 

It is essential to recognise that neither sex nor gender are binary \cite{sumerau2020tale}. Intersex people have chromosomes or bodies that do not fit clearly into the constructed categories of \textit{male} or \textit{female} \cite{ainsworth2015sex}, while gender can be described as a spectrum encompassing masculinity, femininity, androgyny, and everything in between. In technology, this complexity informs the design of UIs like that of Meta's Facebook, which, since 2014, has offered 58 options for users to identify their gender \cite{bivens2017gender}.

While gender is socially constructed, so is gender inequality, which has measurable effects. Women experience bias and discrimination in many contexts, including access to education and fair pay in employment, are underrepresented in political life, and are disproportionately at risk of gender-based violence \cite{unesco2023gender}.
In scientific research, for this and other reasons, it is often important to include participants' sex and/or gender in our analyses \cite{tannenbaum2019sex}. Researchers can decide which is relevant depending on whether the object of study is related to biological or social and psychological characteristics. In the case of speech, both are important. For example, people who have experienced testosterone puberty tend to have longer and thicker vocal cords, producing lower frequencies in their speech on average \cite{hillenbrand1995acoustic}. Illustrating how reductive it is to generalise about \textit{male} or \textit{men's} voices, this group includes most cisgender men, some trans women, many trans men, and many non-binary people \cite{zimman2017gender, tripp2022perceiving}. 

\section{A novel problem?}
We emphasise that understanding sex and gender beyond the binary is not a new idea. While it is true that language, terminology, and legal recognition change over time, it would be wrong to characterise the existence of intersex, trans, non-binary, and gender non-conforming people as a new phenomenon. People who we would now describe as transgender have existed in every recorded human civilisation, from ancient legends to the present day \cite{feinberg1997transgender}. In the context of Western medicine, gender affirming hormone therapy and surgeries have been documented since the 1920s and 1930s \cite{crocq2021gender}. Speech technology, meanwhile, is a comparatively young field. There is no reasonable argument for ignoring the existence of trans and non-binary identities or the ethical considerations that relate to them.

\section{Methods}
We review publications at ISCA Interspeech conferences to interpret how different publications use the variables of gender and sex in their analyses, and what they mean with those terms. We filter papers available at the ISCA archive by venue (Interspeech), year (2013-2023) and keywords (\textit{gender} and \textit{sex}). We further filter out papers that were flagged by the search engine but (1) are not related to `gender' and/or `sex', or (2) papers that only mention gender/sex to provide information about their data being matched by those characteristics, but they are not incorporated as part of their analysis. After applying this filtering, we review 107 papers in total\def\thefootnote{2}\footnote{List of papers available at: https://github.com/ariadnasc/gender-in-speech-research}. We notice that, although the average of Interspeech accepted papers from 2015 to 2018 is 767.5 papers per year\def\thefootnote{3}\footnote{https://www.openresearch.org/wiki/INTERSPEECH}, only 107 from the last 10 years relate to this topic. Therefore, it appears that research published on how speech, sex and gender interact is scarce at this venue. 

We read each paper in full, focusing on (1) whether gender/sex is terminology that the authors define, or its meaning is implied through reading; (2) whether their definition of gender/sex goes beyond the binary male/female or man/woman; (3) whether they differentiate gender from perceived gender, for example in cases where listeners guess someone's gender based on their speech characteristics, which may differ from their actual self-identified gender \cite{weirich2018gender}; and (4) whether the authors acknowledge sex and gender separately, and how each one has a role in the speech characteristics of a speaker. 

\section{Results}

Table \ref{tab:main_results} summarises the main findings of the review conducted. From the total of 107 papers reviewed, 73.8\% of them do not define the term \textit{gender}, leaving its interpretation to the reader. Moreover, 86\% of the papers portray a binarised definition of gender, mostly caused by the lack of diversity in the datasets used (e.g. \cite{burkhardt2010database}). In 82\% of the papers, the term \textit{perceived gender} is not used or defined. However, we observe that a small number of papers evaluate perceived gender through human listening tests (e.g. \cite{hope20_interspeech}). The terms \textit{gender} and \textit{sex} are not differentiated in 80.4\% of the reviewed papers, often using the term \textit{gender} as an umbrella term or using them interchangeably (e.g. \cite{ananthapadmanabha18_interspeech}).

\begin{table*}[]
\centering\begin{tabular}{lcccc}
                                                  & Yes & No  & Partially & Unclear \\ \hline
Is a definition for gender/sex provided?               & 12  & 79 & 16        & N/A     \\
Does the definition of gender/sex go beyond the binary?& 9 & 92  & N/A       & 6      \\
Is perceived gender mentioned?                  & 16  & 88  & 3         & N/A     \\
Do they differentiate between sex and gender? & 14  & 86  & N/A       & 7       \\ \hline
\end{tabular}
\caption{Summary of results for the questions presented in the Methods section. For papers where the question was not clearly answered, we define it as \textit{Unclear}, and for those whose answer is partially or it is answered with related terms, we define it as \textit{Partially}.}
\label{tab:main_results}
\end{table*}

\begin{figure}[t]
  \centering
  \includegraphics[width=\linewidth]{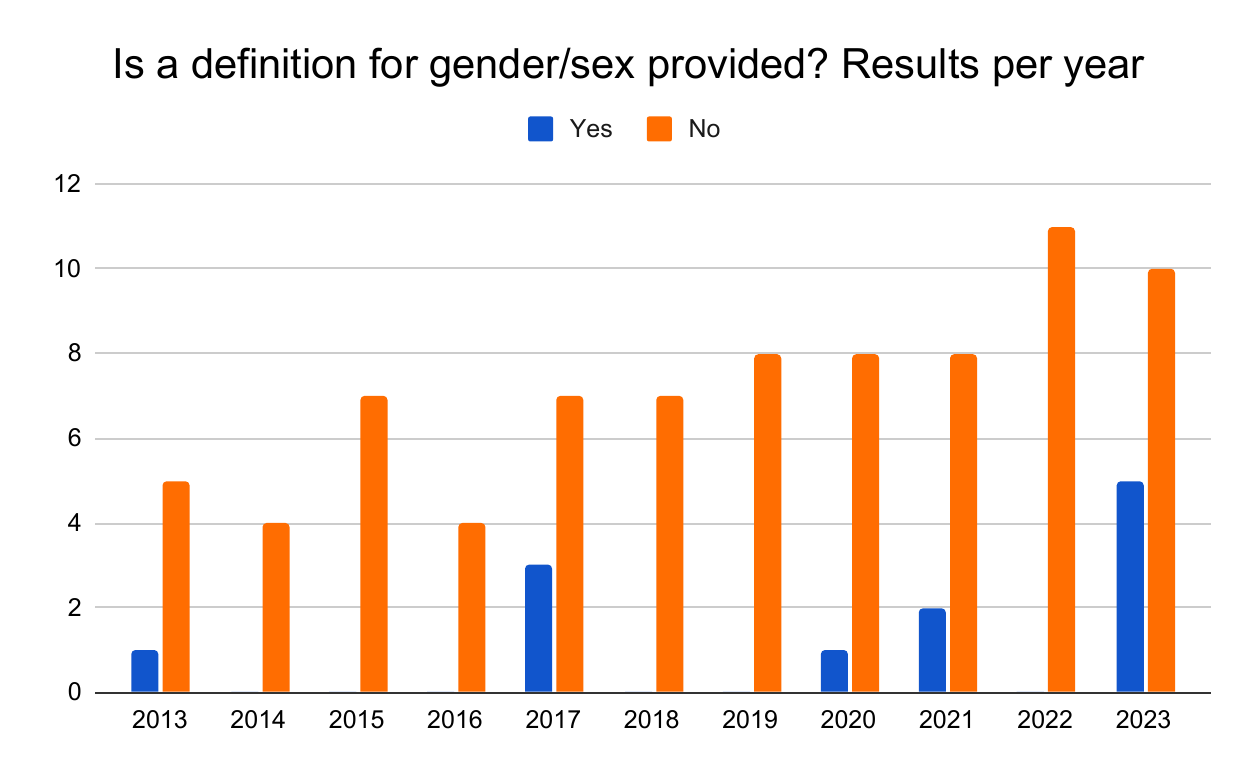}
  \caption{Breakdown of the question \textbf{Is a definition for gender/sex provided?}, per year.}
  \label{fig:binary_per_year}
\end{figure}

\begin{figure}[t]
  \centering
  \includegraphics[width=\linewidth]{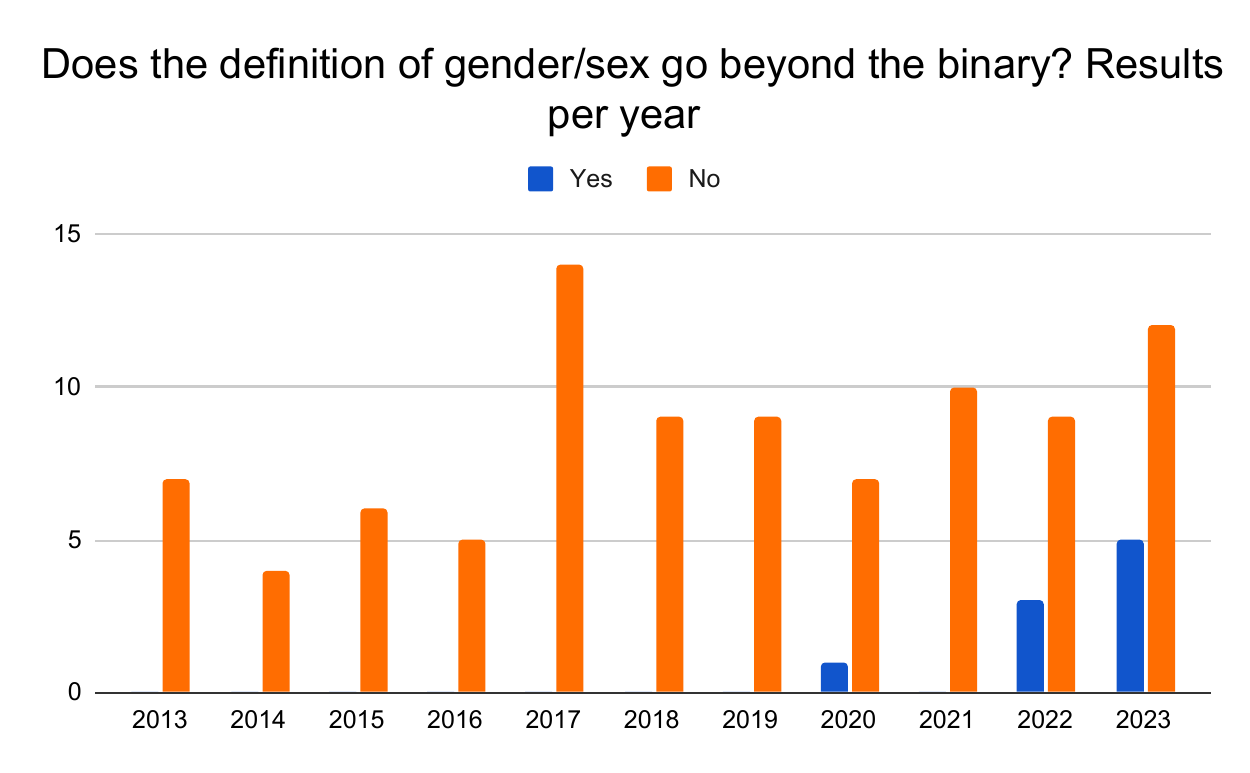}
  \caption{Breakdown of the question \textbf{Does the definition of gender/sex go beyond the binary?}, per year.}
  \label{fig:beyond_binary_per_year}
\end{figure}

While the publications reviewed span 10 years, we notice that more publications in the last 2-3 years include a definition of the term gender or sex, with some of them also defining it beyond binary categories. To quantify this, we plot the ratio of papers with the term gender/sex defined vs. undefined per year (Figure \ref{fig:binary_per_year}), and the ratio of publications with a binarised vs. non-binarised gender/sex description per year (Figure \ref{fig:beyond_binary_per_year}). We observe the beginning of an upward trend for both reviewed questions from 2023. However, even for 2023, the amount of papers that do not define gender/sex and/or use binary definitions of them still dominate.

\section{The impact of words}
When writing our work, we may be tempted to use certain terminology that we assume to be universal and understood by readers. Based on our review, \textit{gender} appears to be such a term, with 73.8\% of publications reviewed not explicitly defining the term. The lack of definition of a term can lead to confusion, more so if the interpretation of such term differs between writer and reader. As defined earlier, we regard gender as a social construct, and different from both perceived gender (what a listener could perceptually guess), and sex (biological attributes of the speaker). 

However, we observe that the term \textit{gender} might not be as easily interpretable without defining what the term actually means. Even though most publications reviewed define \textit{gender} implicitly, we find multiple uses for the term. Some papers use \textit{gender} as an umbrella term for behaviour, personality traits, and physiological properties of someone's voice (\textit{gender} and \textit{sex}). Other work uses the term \textit{gender} to discuss physiological properties (\textit{sex}), e.g. \cite{kabil18_interspeech}, or uses both interchangeably, e.g. \cite{shepstone13_interspeech, altrov13_interspeech}.  
Encouragingly, some publications clearly state what gender is (and most importantly, what it is not), e.g. \cite{hoffmann2021even, szekely23_interspeech, arts23_interspeech}, while others also include what limitations their work might have \cite{zanonboito22_interspeech}. 

The fact that a term is being used with different implicit meanings can be detrimental for research progress, particularly for progress that is inclusive. This can be the case even when the solutions proposed aim at ``diversity" \cite{hoffmann2021even}. Research has the power to improve our society, but also to exacerbate discrimination against certain minority groups, in this case non-binary, trans, agender, gender-fluid, intersex, or other identities. We owe them to be transparent on how we define terms, and acknowledge the limitations of our work, and where those limitations come from. For speech applications, we normally focus on three main terms that are inherently different; (1) \textit{gender}, as defined above, (2) \textit{sex}, as also defined above, and (3) \textit{perceived gender}, which refers to a listener's perception of someone's gender based on their speech characteristics \cite{weirich2018gender}. Therefore, these terms should not be used as umbrella terms or interchangeably.

\section{Inclusive datasets?}
Several of the papers we review were limited to treat gender as binary because it was coded that way in the datasets available. Speech or other data in datasets can be labelled on the basis of participants' self-definition, or by third party annotators. Most available datasets are only composed of binary labels (male/female), e.g. LibriSpeech \cite{panayotov2015librispeech} for automatic speech recognition, IEMOCAP \cite{busso2008iemocap} for speech emotion recognition, or aGender \cite{burkhardt2010database} for age and gender recognition. Other commonly used datasets for speech, such as CommonVoice \cite{ardila2019common}, include a small percentage of voices where voice donors identify as trans, non-binary or intersex. The use of datasets can limit not only the type of research that we can do, but also its scope for inclusivity. Acknowledging the limitations of such datasets can be a good starting point, as in \cite{zanonboito22_interspeech}. 

Datasets are not only limited in terms of how many gender/sex categories may be available, but also in how the dataset was annotated \cite{scheuerman2021datasets}. Some datasets may have used automatic gender recognition (AGR) or human annotation to assign gender labels to speakers, e.g. \cite{kumar16_interspeech}. Such labels would not necessarily correspond to someone's gender, as they would only be guesses based on a model or on a listener's perspective. In cases where gender information cannot be collected from participants, datasets should be specified differently and state that the labels correspond to perceptual gender, as proposed in \cite{markl2022mind}.

Lack of diversity in current datasets is not the only issue towards more inclusive datasets. Crowd work platforms such as Prolific and MTurk force participants to choose a binary category (male/female), in the case of Prolific that of their official document. However, in the case of official documents, countries such as Australia, New Zealand, Germany, India, etc., allow for a third gender marker, X -- therefore, the exclusion of options beyond the binary does not cover all possible options in official documents worldwide. Currently, Prolific includes the option to add gender identity and trans history information on top of that of their official document, which allows for more expressiveness of one's self. Additionally, data privacy is a concern for minority groups, and should be considered when designing experiments \cite{tomasev2021fairness}. However, having to choose from a binary category (in the case of sex/official documents) can not only negatively affect, or even exclude, potential participants \cite{cameron2019gender}, but also hinder the possibility of obtaining a varied sample. This is not only a problem when creating new datasets or evaluating our models, but may also exclude people from participating in research studies.

\section{Inclusive research?}

\textbf{Automatic gender recognition (AGR)}

Twelve of the 107 papers we reviewed focus on speech-based \textit{automatic gender recognition} (AGR), an area where it is especially problematic to operationalise gender as binary and immutable. As \cite{keyes2018misgendering} points out in the context of visual AGR, and as we note above, it is not possible to genuinely `recognise' a person's gender by inference; what AGR systems do is assign gender, making them error-prone for specific groups. This leads us to question the utility of such systems \cite{hoffmann2021terms}. When systems are trained on oversimplified datasets which exclude trans, non-binary and gender non-conforming individuals, and then deployed in the real world where such people exist, they are liable to cause persistent and predictable harm.  We refer the reader to the discussion of \#TravelingWhileTrans in \cite{costanza2020design} for a vivid account of lived experience of this problem. \\

\noindent\textbf{Synthesising gendered and gender-ambiguous speech} 

In the context of text-to-speech (TTS) systems and virtual agents, the human tendency to anthropomorphise interactive agents, including ascribing gender to them, has been a much discussed topic in recent years due to concerns about voice assistants like Siri and Alexa being presented as stereotypically female by default \cite{west2019d}. Researchers and consumers voiced concerns that habitual use of these systems could magnify existing social biases against real women, and evidence of harsher, more impersonal treatment of female subordinate staff in workplaces following `female' voice assistant usage has been reported \cite{weisman2020instantaneity}. The relevance of this issue is reflected in the works that we reviewed pertaining to TTS; several (of the few) papers on this topic addressed the problem of producing gender-neutral synthesised speech, which could be a promising approach to the problem of misogynistic attitudes in this context. However, looking beyond this venue, much progress clearly remains to be made in tackling binary classification and addressing stereotyping in TTS and related work. We have noted statements like `male voices [...] are associated with higher competence' \cite{dou2022effect}, presented uncritically -- despite a large body of work showing that negative evaluations of speech often express social biases (e.g. \cite{levon2021accent}). \\\\

\noindent\textbf{Gender differences in speech and speech perception}

In this research area, some of the work we reviewed was more nuanced. For example, the authors of \cite{barbier15_interspeech} acknowledge that both anatomical factors and a speaker's performance affect the perception of `sexual dimorphism' in humans, but do not discuss the possibility of any mismatch between birth-assigned sex and gender. However, the majority of papers make no distinction at all between biological and social characteristics, leading to unclear statements such as the claim that vocal tract length (VTL) `depends on the speaker’s gender' \cite{ananthapadmanabha18_interspeech}. We also reviewed some papers focusing on gender-affirming therapeutic or surgical interventions for transgender people, noting that here, too, non-binary identities are not mentioned and that there are some problems with the way the authors describe participants. For example, in \cite{liebig22_interspeech} the subjects are reported as `34 male, 35 female, 17 transgender' speakers -- wording which conflicts with the social awareness that trans women are women.

\section{Who are we building systems for?}

As our review shows, sex and gender are both underspecified and inaccurately described as binary in the majority of related research at ISCA Interspeech in the last decade. It is normal and expected that models contain assumptions and simplifications of the real world, and that datasets have `gaps'. We encourage the research community to be mindful of the fact that these assumptions and exclusions are never neutral: the choices we make about what to ignore or leave out of our data, and the categories we use, will inevitably reinforce existing social hierarchies and biases if we do not actively work to challenge them \cite{markl2022mind}. In the case of speech technology, the work that we have reviewed appears to over-represent and serve cisgender\def\thefootnote{4}\footnote{those whose current gender identity aligns with their birth-assigned sex category.}, endosex\def\thefootnote{5}\footnote{those whose innate physical sex characteristics align with cultural expectations of typical male or female bodies.} people, thereby replicating the social problem that trans, non-binary, and intersex people are ignored and marginalised in countless contexts \cite{keyes2018misgendering}.

Many of the papers we reviewed do not state the motivations behind their work or discuss its possible use cases and impacts. Innovations such as an algorithm for `voice gender conversion' (e.g. \cite{sanchez14_interspeech}) have the potential to help people who are going through transition and gender-affirming speech therapy; however, in the wrong hands, they could also be used in unethical ways. As the authors do not mention their motivation, the intended impact of the work is unclear. Furthermore, it should be recognised that even when intentions are good, ethical problems can arise due to the marginalisation of specific groups of people who are excluded and overlooked \cite{hoffmann2021even}. We noted authors' benevolent intentions in some of the papers on AGR. 
However, training models to perform well at the task of detecting people's sex assigned at birth is unlikely to be a responsible use of researchers' time and skills, because in the worst case, such a system is essentially a machine for `outing' and discriminating against trans people. This would be a clear case of algorithmic injustice -- creating models that work for some populations while systematically failing for others \cite{keyes2018misgendering}.

\section{Conclusions}
Our review observes that gender-related speech research is a topic with a scarce amount of publications at ISCA Interspeech in the last 10 years. Most of the work reviewed oversimplifies gender and sex categories, without exploring beyond the binary categories of male/female or men/women. This can be problematic for the future of gender-related speech research, as it alienates minority gender groups from the research we conduct and, most importantly, the models that we build.
We propose the following considerations towards more inclusive gender-related research. With them, we do not aim to be prescriptive for two main reasons: (1) each research study is different and, therefore, needs different considerations, and (2) gender theory is a growing and changing field of research.

\noindent\begin{enumerate}[leftmargin=0.5cm]
    \item \textbf{What is your research about?} We need to be specific about what we are writing about, whether it is gender, sex, or perceived gender. We should avoid umbrella terms and expect interpretations by the reader which may differ from those of the authors.
    \item \textbf{What are the limitations of your work?} Our work is bound to have limitations. Stating them clearly acknowledges them and allows for potential future work and innovation in filling those gaps \cite{markl2022mind}.
    \item \textbf{What are your dataset's limitations and gaps?} When using existing datasets, it is important to understand how the data was collected, and what are the limitations that arise from it. We encourage the reader to state those limitations clearly. When creating new datasets, we encourage the reader to think about possible data gaps, as proposed in \cite{markl2022mind}.
    \item \textbf{How are you collecting demographic information?} Collecting demographic information is valuable in speech research, for both datasets and listening tests. How we ask for this information is key for a good experience for participants and a wider range of participation \cite{spiel2021they, guyan2022queer}. As exemplified in \cite{markl2022mind}, the 2021 release of the Common Voice English dataset (7.0) \cite{ardila2019common} only allowed voice donors to provide a binary gender label (``female'', ``male''). Nonetheless, its early 2024 release (17.0) has expanded the options beyond the binary (``female/feminine'', ``male/masculine'', ``transgender'', ``non-binary'', and ``don't wish to say''). Even though providing a fixed range of categories can still exclude those who identify outside these categories \cite{guyan2022queer}, including explicit options outside the binary is a step in the right direction.
    \item \textbf{What are the assumptions of your work?} Our work may be based on different assumptions. For example, in the case of using a binary definition of gender in our research, not stating that this is an oversimplification makes marginalised groups invisible.
    \item \textbf{What is your model for? Who does it benefit? And most importantly, is it harming anyone?} We need to think about the purposes and motivations of our work, and how it ties into today's society. AGR is an example of technology which could have a negative impact in suggested public use cases such as targeted advertising and security; by failing to consider the existence of bodies or voices that do not conform to stereotyped binary norms, AGR developers can leave marginalised people open to embarrassment or distress if they find themselves misgendered by machines \cite{keyes2018misgendering}.
    Conversely, the authors of \cite{szekely23_interspeech} reflect upon the impacts of their gender-ambiguous neural TTS system, noting that it could benefit people who need artificial voices and do not find themselves represented by traditionally masculine or feminine-sounding options; their work is also valuable for research into implicit gender-related bias and stereotyping.
    As the latter example demonstrates, speech technology advancements should not only involve the majority -- they should include everyone.
\end{enumerate}

\section{Acknowledgements}
We would like to thank Eddie Ungless, Jeremy Steffman, Moishe Holleb, Catherine Lai and Simon King for their valuable insights and feedback on this work. This work was supported by the UKRI Centre for Doctoral Training in Natural Language Processing, funded by the UKRI (grant EP/S022481/1).

\bibliographystyle{IEEEbib}
\bibliography{refs}

\end{document}